\documentclass{esannV2}
\usepackage{graphicx} % Necessary for inserting images
\usepackage{float}
\usepackage{amssymb,amsmath,array}
\usepackage[normalem]{ulem} % Add this line in the preamble
\usepackage[colorlinks=true,urlcolor=teal,linkcolor=teal,citecolor=teal]{hyperref}

\date{November 2024}

\usepackage[usenames,dvipsnames]{xcolor}

\begin{document}
% \title{Latent Sequence Encoding for Transferable Policies in Offline Reinforcement Learning}

\title{TEA: Trajectory Encoding Augmentation for Robust and Transferable Policies in Offline Reinforcement Learning}

%***********************************************************************
% AUTHORS INFORMATION AREA
%***********************************************************************
\author{Batıkan Bora Ormancı$^1$$^2$, Phillip Swazinna$^2$, Steffen Udluft$^2$, Thomas A.\ Runkler$^1$$^2$
%
% Optional short acknowledgment: remove next line if non-needed
% \thanks{This is an optional funding source acknowledgement.}
%
%
% DO NOT MODIFY THE FOLLOWING '\vspace' ARGUMENT
\vspace{.3cm}\\
%
%
% Addresses and institutions (remove "1- " in case of a single institution)
1 Technical University of Munich, Germany
%
% Remove the next three lines in case of a single institution
\vspace{.1cm}\\
2 Siemens AG, Munich, Germany 
}
%***********************************************************************
% END OF AUTHORS INFORMATION AREA
%***********************************************************************

\maketitle

\begin{abstract}
In this paper, we investigate offline reinforcement learning (RL) with the goal of training a single robust policy that generalizes effectively across environments with unseen dynamics. We propose a novel approach, Trajectory Encoding Augmentation (TEA), which extends the state space by integrating latent representations of environmental dynamics obtained from sequence encoders, such as AutoEncoders. Our findings show that incorporating these encodings with TEA improves the transferability of a single policy to novel environments with new dynamics, surpassing methods that rely solely on unmodified states. These results indicate that TEA captures critical, environment-specific characteristics, enabling RL agents to generalize effectively across dynamic conditions.
\end{abstract}

\section{Introduction \& Related Work}

Reinforcement Learning (RL) has seen notable success in addressing complex decision-making tasks across different fields. One ongoing challenge is to enable RL agents to generalize effectively when exposed to environments with varying dynamics, especially when the training needs to happen offline. This issue limits RL's practical applications in real-world situations, where conditions may change unexpectedly. While transfer learning in RL usually seeks to tackle this issue by leveraging online data in both source and target environments, an open challenge remains: developing policies to new, unseen dynamics through adjusted offline training, purely with data from the source environments.

Some methods involve training agents across environments with varying dynamics to foster robustness and adaptability. Domain randomization \cite{tobin2017domain} introduces environmental variations during training to improve robustness, facilitating transfer from simulation to reality. Information bottlenecks \cite{lu2020dynamics} can also be used for dynamics generalization by enabling agents to capture only the task-relevant information while ignoring irrelevant environmental variations.

% Meta-RL algorithms, such as PEARL \cite{rakelly2019efficient} and VariBAD \cite{zintgraf2020varibad}, learn latent task representations to facilitate rapid adaptation to new tasks through online interactions. However, such methods typically require additional environment interactions or task identifiers, which may not be feasible in offline settings.

Meta-reinforcement learning (Meta-RL) algorithms, such as Probabilistic Embeddings for Actor-Critic Reinforcement Learning (PEARL) \cite{rakelly2019efficient} and Variational Bayes-Adaptive Deep RL (VariBAD) \cite{zintgraf2020varibad}, learn latent task representations to facilitate rapid adaptation to new tasks through online interactions. A Gaussian Process-based approach has been proposed to transfer knowledge across 
systems with slight variations by modeling correlations between members, focusing on 
multi-task settings \cite{verstraeten2019fleet}. In contrast, our method employs sequence 
encoders to capture latent dynamics, ensuring generalization in offline reinforcement 
learning without requiring task-specific identifiers.

% \textcolor{red}{In addition, Verstraeten et al. \cite{verstraeten2019fleet} proposed a coregionalization-based Gaussian Process framework to transfer dynamics knowledge across similar systems in reinforcement learning, such as variations of the cartpole environment. While their approach leverages Gaussian processes to model system-specific correlations, our method focuses on integrating latent dynamics representations derived from sequence encoders. This distinction allows TEA to operate effectively in offline settings, addressing dynamics generalization without requiring task-specific identifiers or additional environment interactions.}

Sequence encoders, like AutoEncoders and recurrent neural networks, have been used to compress sequences into compact representations in RL. For example, the world models framework \cite{ha2018recurrent} learns latent representations of the environment to facilitate planning.

% In offline RL, where learning is confined to static datasets, algorithms like Batch-Constrained Q-learning (BCQ) \cite{fujimoto2019off,fujimoto2019benchmarking} and LION \cite{swazinna2023userinteractive} address challenges such as value function overestimation and out-of-distribution actions. However, these techniques do not directly tackle the transferability of policies to new environments with unobserved dynamics.
% alternative: unexplored (instead of unobserved) in the last sentence
% LION \cite{swazinna2023we}, while providing user-interactive control over policy conservatism and performance, also operates within a single known environment and does not address the transferability of policies to environments with unobserved dynamics."

In offline RL, where learning is confined to static datasets, algorithms like Batch-Constrained Q-learning (BCQ) \cite{fujimoto2019off,fujimoto2019benchmarking} and Learning in Interactive Offline eNvironments (LION) \cite{swazinna2023userinteractive} address challenges such as value function overestimation and out-of-distribution actions. However, these techniques do not directly tackle the transferability of policies to new environments with unobserved dynamics.

In this study, we investigate an offline method in RL to enhance robustness and generalization, concentrating on the classic CartPole environment with variations in pole lengths and cart masses. We introduce a sequence encoder to capture latent representations of environment dynamics from state-action sequences. By adding these latent encodings to the state space, we aim to improve the agent's adaptability to different environments. Our work provides three main contributions:

\begin{enumerate}
    \item We introduce the use of sequence encoders, particularly AutoEncoders, to derive meaningful representations of environment dynamics from state-action sequences.
    \item We demonstrate that adding these encodings to the state space enhances the transferability of policies to environments with new dynamics.

    \item  We compare our approach to a baseline method and demonstrate that our method enables a single policy to maintain strong performance across new environments after being trained solely on the source environments.
    % alternative phrasing: 'even when the agent hasn't been trained on the new environment' instead of 'as training progresses in the source envirnoments' 
\end{enumerate}

\section{Experimental Setup}
\begin{figure}[H]
    \centering
    \includegraphics[width=1\textwidth]{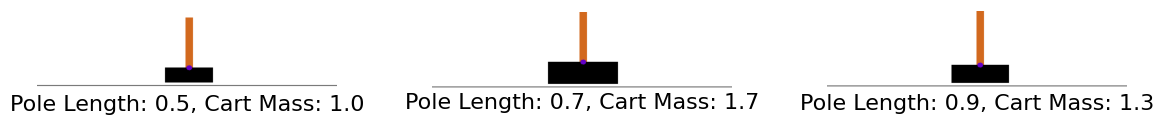}
    \caption{Illustration of cartpoles with varying pole lengths and cart masses.}
    \label{fig:cartpole_images}
\end{figure}

\textbf{Environment Design:} We use the classic CartPole-v1 environment \cite{barto1983neuronlike} as our testbed, introducing variability in dynamics by modifying its transition-defining parameters. Five source environments were created by altering pole length and cart mass, with values drawn uniformly from the ranges $[0.1, 2.0]$ and $[0.5, 2.0]$, respectively. These parameters are set for each environment as follows:

\begin{equation}
    \text{env\_properties} = 
    \left\{ 
    \begin{aligned}
        &\text{pole length}: \quad l_i \sim U(0.1, 2.0), \\
        &\text{cart mass}: \quad m_i \sim U(0.5, 2.0)
    \end{aligned}
    \right\}.
\end{equation}

\begin{figure}[h]
    \centering
    \begin{minipage}[c]{0.5\textwidth} % Adjust the width here as needed
        \includegraphics[width=\textwidth]{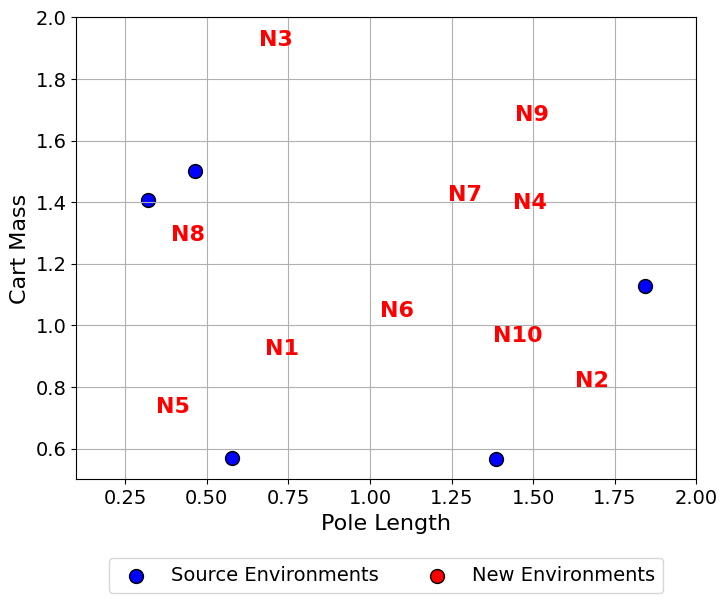}
    \end{minipage}%
    \begin{minipage}[c]{0.4\textwidth} % Adjust caption width
        \caption{Scatter plot showing pole length and cart mass for source environments (blue) and new environments (red).}
        \label{fig:scatter_envs}
    \end{minipage}
\end{figure}

The goal is to train a single policy using data from the source environments that can generalize to new environments without additional training. Our method is designed to achieve robust generalization through offline training on the source environments. Figure \ref{fig:scatter_envs} displays the variability in pole length and cart mass across source environments and highlights the individual new environments N1 to N10 used for evaluation.

% \vspace{\baselineskip}

\textbf{Sequence Encoder:} To capture the dynamics of each environment, we introduce the notion of a sequence encoder. For this, we used AutoEncoders (AE), however other architectures, like LSTMs or even MLPs are conceivable. The AE is trained to reconstruct sequences of state-action pairs, compressing the sequence into a latent representation. The latent encoding is derived from the bottleneck layer of the AE.

% \textbf{Sequence Encoder:} To capture the dynamics of each environment, we introduce the notion of a sequence encoder. For this, we experimented with utilizing AutoEncoders (AEs), Multi-Layer Perceptrons (MLPs), and Long Short-Term Memory networks (LSTMs). The AE is trained to reconstruct sequences of state-action pairs, compressing the sequence into a latent representation. In the MLP and LSTM, the goal is to predict the source environment based on the sequence. The latent encoding is derived from the bottleneck layer of the AE or the hidden state of the MLP/LSTM.

Our experiments indicated that the AE with a four-dimensional latent space produced the best performance, so we focus our analysis on this configuration.

\vspace{\baselineskip}

\textbf{Data Collection:} We initially train a Deep Q-Network (DQN) \cite{mnih2015human} on the standard CartPole-v1 environment, using an implementation from the authors of the BCQ algorithm \cite{fujimoto2019off}. This trained agent is then deployed on a number of source environments with randomly sampled dynamics, as visualized in Figure \ref{fig:scatter_envs}.

% ...  deployed on a multi-environment wrapper, which switches among the five source environments at the start of each episode.

We collect interaction data from this agent, storing transitions in a regular replay buffer. For the AutoEncoder training, we use sequences of state-action pairs of length 16 (comprising 16 states and 15 actions in between), generated by the DQN agent trained on the regular CartPole environment.

\vspace{\baselineskip}

\textbf{Sequence Encoder Training:} The AE is trained to reconstruct input sequences, compressing each sequence into a four-dimensional latent space. The training objective is to minimize the mean squared error between input and reconstructed sequences.
\vspace{\baselineskip}

\textbf{State Space Augmentation:} After training the encoder, the state space of each source environment is augmented. For each source environment in our system that we used for training, five trajectories of length 16 are taken from the buffer and encoded. The average latent encoding is then computed, creating a four-dimensional vector that reflects the dynamics of each environment.

This encoding is appended to the original states, expanding the state space. To avoid additional data collection from the source environments, we retroactively extend states in the existing dataset by appending the relevant encodings.

\vspace{\baselineskip}

\textbf{Offline RL Training with BCQ:} We employ the BCQ algorithm in its discrete form to train policies exclusively using pre-collected datasets from the source environments, without any additional interactions with the environments during training. We conduct the following experiments:

\begin{enumerate}
    \item \textbf{Baseline}: BCQ trained on the data from the source environments without augmented states.
    \item \textbf{Trajectory Encoding Augmentation (TEA)}: BCQ trained on the data from the source environments with augmented states using learned encodings of trajectories.
\end{enumerate}

\textbf{Evaluation Protocol:} 
To assess policy transferability, we measure each policy's performance on new environment(s) which were randomly sampled from the same distribution as the source environments and are visualized in Figure \ref{fig:scatter_envs}, reporting average returns as well as standard error over 100 random seeds after 20,000 timesteps of training. Note that five sequences of length 16 were needed to be seen from these new environments to facilitate the creation of a first encoding.

%\section{Discussion \& Conclusion}
\section{Discussion}

\begin{figure}[H]
    \centering
    \begin{minipage}[c]{0.5\textwidth} % Adjust the width here as needed
        \includegraphics[width=\textwidth]{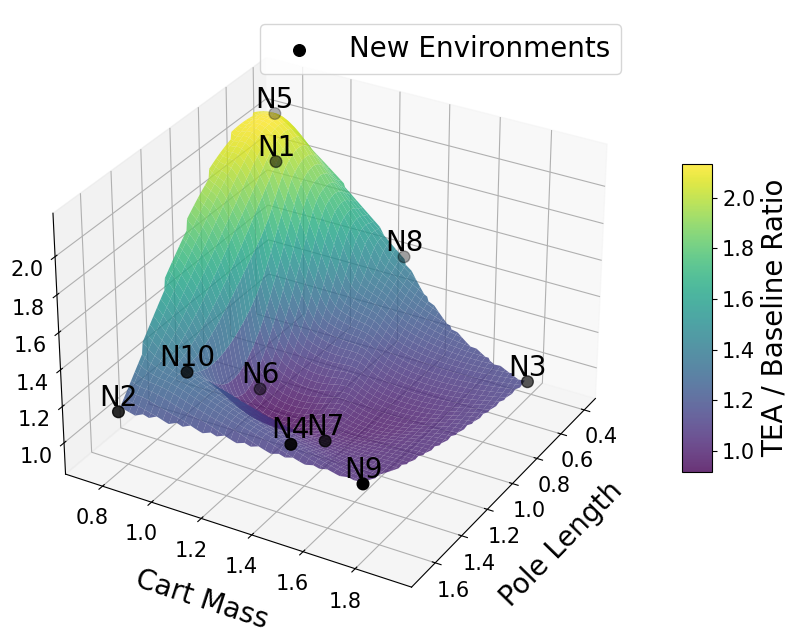}
    \end{minipage}%
    \hspace{0.05\textwidth} % Adjust horizontal space as needed
    \begin{minipage}[c]{0.4\textwidth} % Adjust caption width
        \vspace{0.5cm} % Add vertical space here
        % \caption{Plot illustrating performance increase achieved through the encoding augmentation (TEA) for different new environments, compared to the baseline performance.}

        \caption{Plot illustrating the performance improvement achieved through TEA across various new environments, compared to the baseline performance. Notably, all performance ratios exceed 1, demonstrating consistent gains.}
        \label{fig:perf_encoding_div_baseline}
    \end{minipage}
\end{figure}

\textbf{Results:} The performance of each experiment on the new environment(s) after 20,000 timesteps of training on the source environments is shown in Table \ref{tab:performance_across_environments_prof}. In Figure \ref{fig:perf_encoding_div_baseline} we see the performance of trajectory encoding augmentation (TEA) divided by the baseline performance for each new environment.

\begin{table}
\centering

\begin{tabular}{|l|c|c|} 
\hline
\textbf{Environment} &  \textbf{Baseline} & \textbf{TEA} \\
\hline
N1 & 36.2 ± 2.2 & \textbf{72.1 ± 6.6} \\
N2 & 38.4 ± 0.2 & \textbf{45.3 ± 2.2} \\
N3 & 43.3 ± 1.1 & \textbf{50.1 ± 2.9} \\
N4 & 49.8 ± 0.7 & \textbf{55.7 ± 2.3} \\
N5 & 36.3 ± 2.2 & \textbf{72.5 ± 6.7} \\
N6 & 35.1 ± 0.6 & 35.2 ± 1.5 \\
N7 & 47.3 ± 0.8 & 47.7 ± 1.7 \\
N8 & 101.5 ± 1.4 & \textbf{145.5 ± 7.4} \\
N9 & 58.9 ± 1.9 & 60.7 ± 2.5 \\
N10 & 40.3 ± 0.9 & \textbf{52.6 ± 13.7} \\
\hline
\textbf{Average} & 48.7 ± 0.4 & \textbf{63.7 ± 1.9} \\
\hline
\end{tabular}
\caption{Performance Across N1 to N10 (mean ± SEM over 100 seeds)}
\label{tab:performance_across_environments_prof}
\end{table}

% As training progresses, the performance of the baseline decreases, while TEA sustains a higher performance with continued training on the source environments.

% \vspace{\baselineskip}
\textbf{Analysis:} The results in Table \ref{tab:performance_across_environments_prof} demonstrate a statistically significant improvement in performance for the TEA method over the baseline condition, as indicated by the uncertainties accompanying the average values. These findings show that augmenting the state space with latent encodings, derived from a sequence encoder, enables the trained policy to generalize more effectively across environments with new dynamics. The observed results confirm that the approach based on latent encodings leads to better outcomes under changing dynamics, suggesting that this form of augmentation enhances policy transferability compared to the baseline.

\section{Conclusion}

In this paper, we addressed the problem of enabling quick adaptation to new environments with different dynamics using offline pretraining from source environments. Our proposed Trajectory Encoding Augmentation (TEA) method demonstrates that using sequence encoders to augment the state space in offline RL can notably improve policy transferability to respective new environments. By utilizing latent representations of environment-trajectories, agents perform well in new environments.

This technique offers a promising path toward developing RL agents more resilient to changing environmental dynamics. 

\textbf{Future Work:} Our method can be expanded to explore the \emph{offline-to-online} learning paradigm. Specifically, we plan to initiate training using BCQ and subsequently continue with regular Q-learning. This extension would enable us to examine the agent’s adaptability during the shift from offline to online learning, yielding deeper insights into the effectiveness of our approach and offering a more comprehensive perspective on its application in dynamic environments. In addition, exploring environments with higher-dimensional state-action spaces and experimenting with alternative encoder architectures could further validate and enhance the versatility of our approach.

% Also, performance can potentially be further improved and use-cases for the method extended through recreation of the encoding through utilizing a sliding window of trajectories as encoder input throughout the interaction with the new environment. 

% Future Work: Building on the promising results of Trajectory Encoding Augmentation (TEA), future research could extend this approach in several directions. First, integrating TEA into an offline-to-online learning paradigm would allow initial training with BCQ to transition seamlessly into online reinforcement learning algorithms, such as regular Q-learning. This hybrid strategy would enable the agent to adapt dynamically to previously unseen environments, leveraging both the robustness gained during offline training and the adaptability of online fine-tuning.

% Moreover, exploring environments with higher dimensional state-action spaces, such as robotic control tasks or multi-agent systems, would provide a broader validation of the method's applicability. Employing advanced sequence encoder architectures, including Transformer-based models or variational approaches, could further enhance the richness and utility of the learned latent representations. Finally, real-time encoding of trajectories during interaction with new environments—via a sliding window approach—could enable continual refinement of the state augmentation, improving adaptability to dynamic or non-stationary environments. These avenues offer exciting opportunities to refine and generalize the TEA framework for diverse real-world applications.

\begin{footnotesize}

\bibliographystyle{unsrt}
% \bibliography{references-shortened}
\bibliography{references}

\end{footnotesize}

\end{document}